\documentclass[runningheads]{llncs}
\usepackage[T1]{fontenc}

\usepackage{graphicx,verbatim}
\usepackage{amsmath}
\usepackage{subcaption}
\usepackage{float}
\usepackage{xcolor}
\usepackage{url}
\usepackage{subcaption}
\usepackage{caption}
\usepackage{multirow}

\begin{document}
\title{Advancing Fetal Ultrasound Image Quality Assessment in Low-Resource Settings}

\author{Dongli He \and Hu Wang \and Mohammad Yaqub}
\authorrunning{D. He et al.}
\institute{Mohamed bin Zayed University of Artificial Intelligence, Abu Dhabi, UAE\\
\email{\{dongli.he, hu.wang, mohammad.yaqub\}@mbzuai.ac.ae}}

\maketitle              
\begin{abstract}
Accurate fetal biometric measurements, such as abdominal circumference, play a vital role in prenatal care. However, obtaining high-quality ultrasound images for these measurements heavily depends on the expertise of sonographers, posing a significant challenge in low-income countries due to the scarcity of trained personnel. To address this issue, we leverage FetalCLIP, a vision-language model pretrained on a curated dataset of over 210,000 fetal ultrasound image-caption pairs, to perform automated fetal ultrasound image quality assessment (IQA) on blind-sweep ultrasound data. We introduce FetalCLIP\textsubscript{CLS}, an IQA model adapted from FetalCLIP using Low-Rank Adaptation (LoRA), and evaluate it on the ACOUSLIC-AI dataset against six CNN and Transformer baselines. FetalCLIP\textsubscript{CLS} achieves the highest F1 score of 0.757. Moreover, we show that an adapted segmentation model, when repurposed for classification, further improves performance, achieving an F1 score of 0.771. Our work demonstrates how parameter-efficient fine-tuning of fetal ultrasound foundation models can enable task-specific adaptations, advancing prenatal care in resource-limited settings. The experimental code is available at: \url{https://github.com/donglihe-hub/FetalCLIP-IQA}.


\end{abstract}
\section{Introduction}

Ultrasound has long played a vital role in routine prenatal care~\cite{Whitworth2010}. Fetal biometric measurements, such as abdominal circumference (AC), biparietal diameter (BPD), and femur length (FL), are essential for assessing fetal growth and monitoring high-risk pregnancies. However, acquiring accurate measurements requires considerable expertise from sonographers to identify appropriate standard planes. In regions where experienced personnel are scarce, blind-sweep ultrasound, typically performed by novice operators using low-cost portable probes, is often employed to collect obstetric scans~\cite{Self2022CALOPUS}. These blind-sweep scans usually yield low-quality ultrasound data that may lack the precise anatomical planes conventionally required for accurate biometric assessment.

Recent advances in machine learning have demonstrated remarkable capabilities in medical domains. One important application is automated assessment of fetal ultrasound image quality~\cite{7875138}, which is particularly useful for free-hand ultrasound sequences acquired through blind-sweep protocols as it enables accurate biometric measurements even in the absence of skilled sonographers.

To develop robust machine learning models, a common strategy is to train them on data collected from the same regions where they will be deployed. However, real-world deployment of medical models in resource-constrained settings is often limited by scarce data and computational resources. An effective alternative is to leverage large pretrained models, commonly referred to as foundation models, which are trained on extensive datasets and subsequently adapted to specific downstream tasks through transfer learning.

In this work, we build our models upon FetalCLIP~\cite{maani2025fetalclipvisuallanguagefoundationmodel}, a vision-language foundation model specifically trained for fetal ultrasound, to address the task of fetal ultrasound image quality assessment (IQA). Our objective is to assist less-experienced sonographers in identifying high-quality frames for abdominal circumference measurement. To enable efficient model adaptation, we employ parameter-efficient fine-tuning using Low-Rank Adaptation (LoRA)~\cite{hu2021loralowrankadaptationlarge}, which allows efficient task-specific tuning with minimal trainable parameters. This approach is well-suited for deployment in settings with limited computational resources. Our main contributions are as follows:

\begin{itemize}
    \item We propose \mbox{FetalCLIP\textsubscript{CLS}}, a model adapted from the fetal ultrasound foundation model FetalCLIP using LoRA. FetalCLIP\textsubscript{CLS} consistently outperforms strong CNN and Transformer baselines on the fetal ultrasound IQA task, while requiring only a small number of trainable parameters.
    \item We demonstrate the feasibility of repurposing a segmentation model for IQA tasks. This adapted segmentation model, named FetalCLIP\textsubscript{SEG}, further improves classification performance, achieving a higher F1 score and recall compared to classification approaches.
\end{itemize}

\section{Related Work}

Methods for image quality assessment (IQA)~\cite{ma2025surveyimagequalityassessment} can be broadly classified into two categories: statistical approaches and machine learning-based techniques. Statistical methods include human visual system-based metrics such as the Structural Similarity Index (SSIM)~\cite{1284395}, transform domain-based techniques like BLIINDS-II~\cite{6172573}, and natural scene statistics-based approaches such as the Information Fidelity Criterion (IFC)~\cite{1576816}.

Machine learning-based IQA methods can be categorized into traditional machine learning, convolutional neural network (CNN)-based, and Transformer-based approaches. Traditional machine learning methods such as BRISQUE~\cite{6272356} rely on hand-crafted features derived from natural scene statistics to estimate image quality. CNN-based methods such as IQA-CNN~\cite{6909620} extract deep features from image data to improve prediction accuracy. More recently, Transformer-based methods like TRIQ~\cite{yang2022maniqamultidimensionattentionnetwork} have been proposed to address the locality bias inherent in CNNs by capturing long-range dependencies, leading to improved performance on IQA tasks.

State-of-the-art fetal ultrasound IQA methods predominantly rely on machine learning. Wu et al.~\cite{7875138} employ two CNNs to assess ultrasound scans of the fetal abdominal region, demonstrating performance comparable to subjective ratings from medical experts. Cengiz et al.~\cite{cengiz2023fusqafetalultrasoundsegmentation} propose an automatic method to evaluate the quality of predicted segmentation masks. Boumeridja et al.~\cite{boumeridja2025enhancing} introduce a super-resolution technique that enhances ultrasound image resolution to improve downstream classification performance in low-resource settings.

The majority of models for fetal ultrasound IQA have adopted pretrained models due to their superior performance~\cite{sendrabalcells2023generalisabilityfetalultrasounddeep}. However, most publicly available foundation models are pretrained on natural images, which may not be optimal for domain-specific tasks. FetalCLIP is a foundation model pretrained on fetal ultrasound image-caption pairs using Contrastive Language-Image Pretraining (CLIP)~\cite{radford2021learningtransferablevisualmodels}. We are interested in evaluating its transferability in low-resource settings.

\begin{figure}[h]
    \centering
    \includegraphics[width=0.96\textwidth]{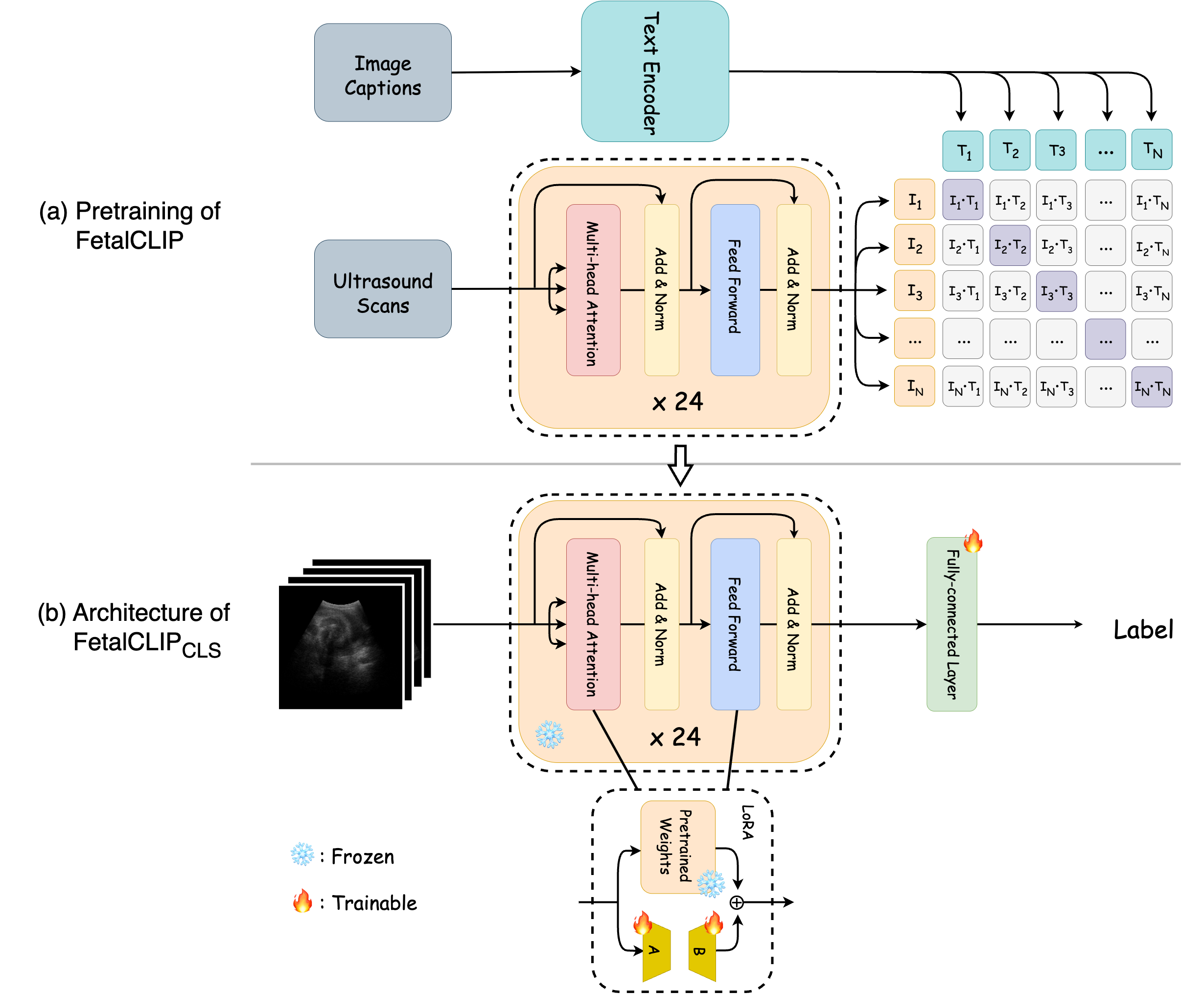}
    \caption{
    (a) Contrastive pretraining of FetalCLIP, where the model learns to align ultrasound scans with their corresponding text descriptions. (b) FetalCLIP\textsubscript{CLS} utilizes the pretrained image encoder from FetalCLIP followed by a linear head to predict whether a frame is optimal for fetal biometric measurement. The image encoder remains frozen during FetalCLIP\textsubscript{CLS} training, with only the LoRA modules and linear head being trainable.
    } \label{fig:model_architecture}
\end{figure}

\section{Methodology} \label{methodology}
\subsection{Task Formulation}

We formulate fetal ultrasound image quality assessment (IQA) as a binary classification problem to identify frames containing clear anatomical structures suitable for fetal biometric measurement. Formally, given a 2D ultrasound frame \( X \), the model predicts a binary label \( y \in \{0, 1\} \), where \( y = 1 \) indicates the presence of relevant anatomical structures and \( y = 0 \) indicates their absence.

\subsection{Model Architecture}

We utilize the image encoder from FetalCLIP~\cite{maani2025fetalclipvisuallanguagefoundationmodel}, which was trained on over 210,000 fetal ultrasound image-caption pairs using CLIP~\cite{radford2021learningtransferablevisualmodels}. As illustrated in Figure~\ref{fig:model_architecture}(a), this encoder captures rich semantic and spatial features by aligning ultrasound images with their corresponding textual descriptions during pretraining.

In our approach, we keep the pretrained image encoder frozen and insert LoRA modules within the attention and feed-forward blocks to enable parameter-efficient fine-tuning. Our ablation study in Appendix~\ref{sec:ablation} demonstrates that fine-tuning transformer-based models with LoRA outperforms both linear probing and full-parameter fine-tuning.

Ultrasound scans are passed through the encoder to produce a one-dimensional embedding that encapsulates semantic and structural information about the images. A classification head, consisting of a single fully connected layer, is appended to the image encoder to map the embedding to a binary prediction indicating whether the frame is suitable for fetal biometric measurement. We refer to this classification pipeline as FetalCLIP\textsubscript{CLS}, as illustrated in Figure~\ref{fig:model_architecture}(b).

\subsection{Leveraging Segmentation Model for Classification Task}

Based on the hypothesis that a segmentation model capable of perfectly annotating an image can also classify it without error, we propose FetalCLIP\textsubscript{SEG}, a segmentation model that employs a thresholding strategy to convert predicted segmentation masks into binary classification labels.

This model leverages the same frozen image encoder from FetalCLIP with embedded LoRA modules for parameter-efficient fine-tuning, followed by a lightweight U-shaped network inspired by UNETR~\cite{hatamizadeh2021unetrtransformers3dmedical}. Ground truth masks, as detailed in Section~\ref{sec:dataset}, are used during training to guide the segmentation process.

\section{Experiments}
\subsection{Dataset} \label{sec:dataset}

We use a 2D B-mode fetal ultrasound dataset from the MICCAI 2024 ACOUSLIC-AI Challenge~\cite{SAPPIA2025103640}. The dataset was originally developed for operator-agnostic abdominal circumference measurement in low-income countries. It comprises fetal abdominal ultrasound scans collected from pregnant women between 20 and 32 weeks of gestation in Sierra Leone and Tanzania. The scans were acquired by novice users with only one hour of training using a low-cost portable ultrasound probe, and each scan is accompanied by an expert-annotated mask. Since not all frames contain abdominal structures, some masks may be empty. Each patient undergoes six blind sweeps: three transverse sweeps in the caudocranial direction and three sagittal sweeps from the patient's left to right, with each sweep containing 140 frames.

In the challenge, a total of 300 cases are provided for training. On average, only 2.6 out of every 100 frames contain abdominal structures, resulting in a highly imbalanced label distribution. To mitigate this imbalance, we exclude sweeps without annotated masks. This filtering process yields a more balanced dataset, increasing the average number of frames with clear abdominal planes to 8.6 per 100 scans.

Since the official validation and test sets are not publicly available, we randomly split the 300 cases into 210 for training, 30 for validation, and 60 for testing. To prevent data leakage, all sweeps from the same patient are assigned to the same split. This results in 52,500 training, 8,540 validation, and 16,380 test ultrasound images.

\subsection{Evaluation Metrics}

We evaluate model performance using widely adopted classification metrics that focus on different aspects: overall correctness (accuracy), the ability to identify optimal frames (precision and recall), and the harmonic mean of precision and recall (F1 score).

\subsection{Implementation Details}

During preprocessing, each image is first padded to a square shape. Data augmentation is then applied to the training set, including color jittering, contrast-limited adaptive histogram equalization (CLAHE)~\cite{10.5555/180895.180940}, and affine transformations. We generate two additional augmented versions for each training frame. All images are subsequently resized to $224 \times 224$ pixels.

We train all models using the AdamW optimizer~\cite{loshchilov2019decoupledweightdecayregularization} with a fixed learning rate of $3 \times 10^{-4}$ for 5 epochs. All experiments are conducted on a single NVIDIA RTX 4090 GPU. Each experiment is repeated five times, and the checkpoint with the lowest validation loss is selected for evaluation on the test set.

All classification models are optimized using binary cross-entropy loss, while FetalCLIP\textsubscript{SEG} is trained with the Dice loss~\cite{milletari2016vnetfullyconvolutionalneural}. To convert segmentation outputs into binary labels, we apply a thresholding strategy: if the number of foreground pixels in the predicted mask exceeds 1\% of the image area (approximately 500 pixels), the frame is labeled as $y = 1$; otherwise, it is labeled as $y = 0$.

\subsection{Results}

We compare FetalCLIP\textsubscript{CLS} against six baseline models: three CNN-based models—DenseNet~\cite{huang2018denselyconnectedconvolutionalnetworks}, EfficientNet~\cite{tan2020efficientnetrethinkingmodelscaling}, and VGG~\cite{simonyan2015deepconvolutionalnetworkslargescale}—and three Transformer-based models—Swin Transformer~\cite{liu2021swintransformerhierarchicalvision}, DeiT~\cite{touvron2022deitiiirevengevit}, and Vision Transformer (ViT)~\cite{dosovitskiy2021imageworth16x16words}.

\begin{table}[h]
\caption{
Model performance on fetal ultrasound IQA. We compare FetalCLIP\textsubscript{CLS} against six baseline models. Metrics are reported as mean ± standard deviation across five independent runs. The last column indicates the number of trainable parameters for each model. The best average scores for each metric are \textbf{bolded}, while second-best scores are \underline{underlined}.
} \label{tab:model_performance}
\resizebox{\textwidth}{!}{
\begin{tabular}{|l|l|c|c|c|c|c|}
\hline
\textbf{Architecture} & \textbf{Models} & \textbf{Accuracy}$\uparrow$ & \textbf{F1 Score}$\uparrow$ & \textbf{Precision}$\uparrow$ & \textbf{Recall}$\uparrow$ & \textbf{\# Trainable} \\
\hline
\multirow{3}{*}{CNN} & DenseNet & $0.9516 \pm 0.002$ & $0.7024 \pm 0.028$ & $\underline{0.7805} \pm 0.026$ & $0.6420 \pm 0.059$ & 7.0 M \\
& EfficientNet & $0.9537 \pm 0.004$ & $0.7253 \pm 0.030$ & $0.7725 \pm 0.025$ & $0.6855 \pm 0.053$ & 4.0 M \\
& VGG & $0.9510 \pm 0.002$ & $0.7084 \pm 0.021$ & $0.7580 \pm 0.023$ & $0.6671 \pm 0.048$ & 134 M \\
\hline
\multirow{4}{*}{Transformer} & Swin & $0.9565 \pm 0.003$ & $0.7429 \pm 0.039$ & $\textbf{0.7864} \pm 0.032$ & $0.7113 \pm 0.087$ & 1.7 M \\
& DEIT & $0.9554 \pm 0.001$ & $0.7466 \pm 0.014$ & $0.7619 \pm 0.035$ & $0.7363 \pm 0.059$ & 2.4 M \\
& ViT\textsubscript{400M} & $\underline{0.9560} \pm 0.003$ & $\underline{0.7506} \pm 0.019$ & $0.7657 \pm 0.042$ & $\textbf{0.7417} \pm 0.067$ & 2.4 M \\
& FetalCLIP\textsubscript{CLS} & $\textbf{0.9575} \pm 0.001$ & $\textbf{0.7570} \pm 0.007$ & $0.7782 \pm 0.034$ & $\underline{0.7397} \pm 0.041$ & 2.4 M \\
\hline
\end{tabular}
}
\end{table}

The CNN baselines are pretrained on ImageNet-1K~\cite{5206848}, and we perform full-parameter fine-tuning for these models. For the Transformer baselines, we freeze the encoder and apply LoRA, following an architecture similar to FetalCLIP\textsubscript{CLS}. In fact, the only difference lies in the choice of encoder for Transformer-based models. We choose Transformer baselines with a comparable number of parameters to FetalCLIP\textsubscript{CLS} to ensure a fair comparison. Swin Transformer and DeiT are pretrained on ImageNet-22K, while the ViT baseline is pretrained on WIT-400M image-text pairs using CLIP~\cite{Cherti_2023}.

As shown in Table~\ref{tab:model_performance}, FetalCLIP\textsubscript{CLS} achieves superior performance in accuracy and F1 score compared to all baselines while maintaining competitive precision and recall with a small number of trainable parameters. Note that FetalCLIP is pretrained on over 210,000 fetal ultrasound image-caption pairs. Although the ViT baseline is pretrained on 400M image-text pairs—a dataset more than 1,900 times larger than that used for FetalCLIP—FetalCLIP\textsubscript{CLS} still outperforms ViT\textsubscript{400M} in accuracy, F1 score, and precision. This demonstrates the effectiveness of foundation models' domain-specific pretraining and highlights the advantages of FetalCLIP\textsubscript{CLS} for tasks requiring high-quality fetal ultrasound representations.

We observe that Transformer-based models consistently outperform CNN-based models in accuracy, F1 score, and recall. Although CNNs were once the preferred choice for many machine learning practitioners, they have been surpassed by Transformers for fetal ultrasound IQA tasks, according to our experimental results. As a general recommendation, practitioners seeking superior model performance should favor Transformer-based models over CNNs when developing fetal ultrasound IQA systems.

\begin{figure}[htbp]
    \begin{subfigure}[b]{0.45\textwidth}
        \includegraphics[width=\textwidth]{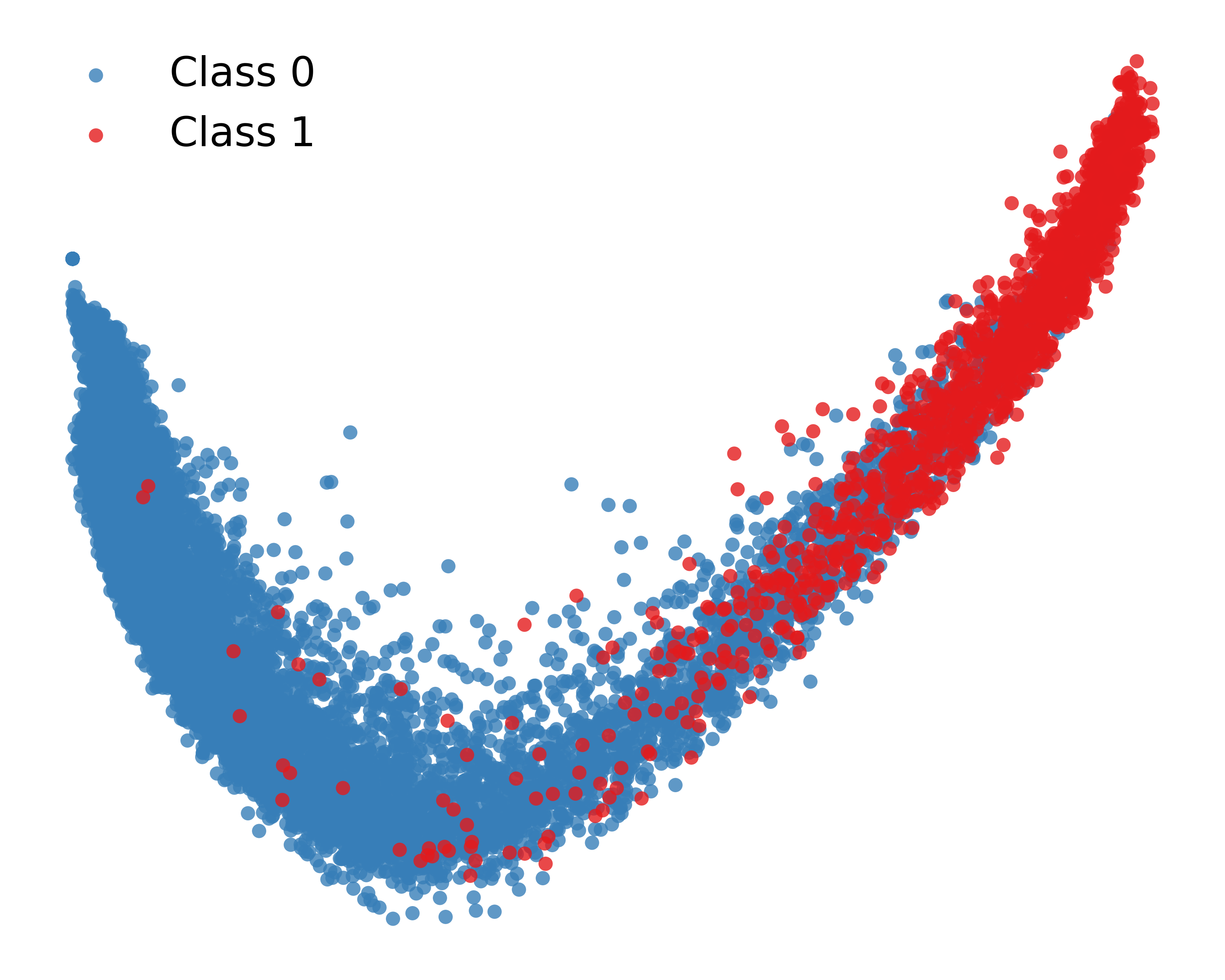}
        \caption{PCA}
        \label{fig:right}
    \end{subfigure}
    \hfill
    \begin{subfigure}[b]{0.45\textwidth}
    \includegraphics[width=\textwidth]{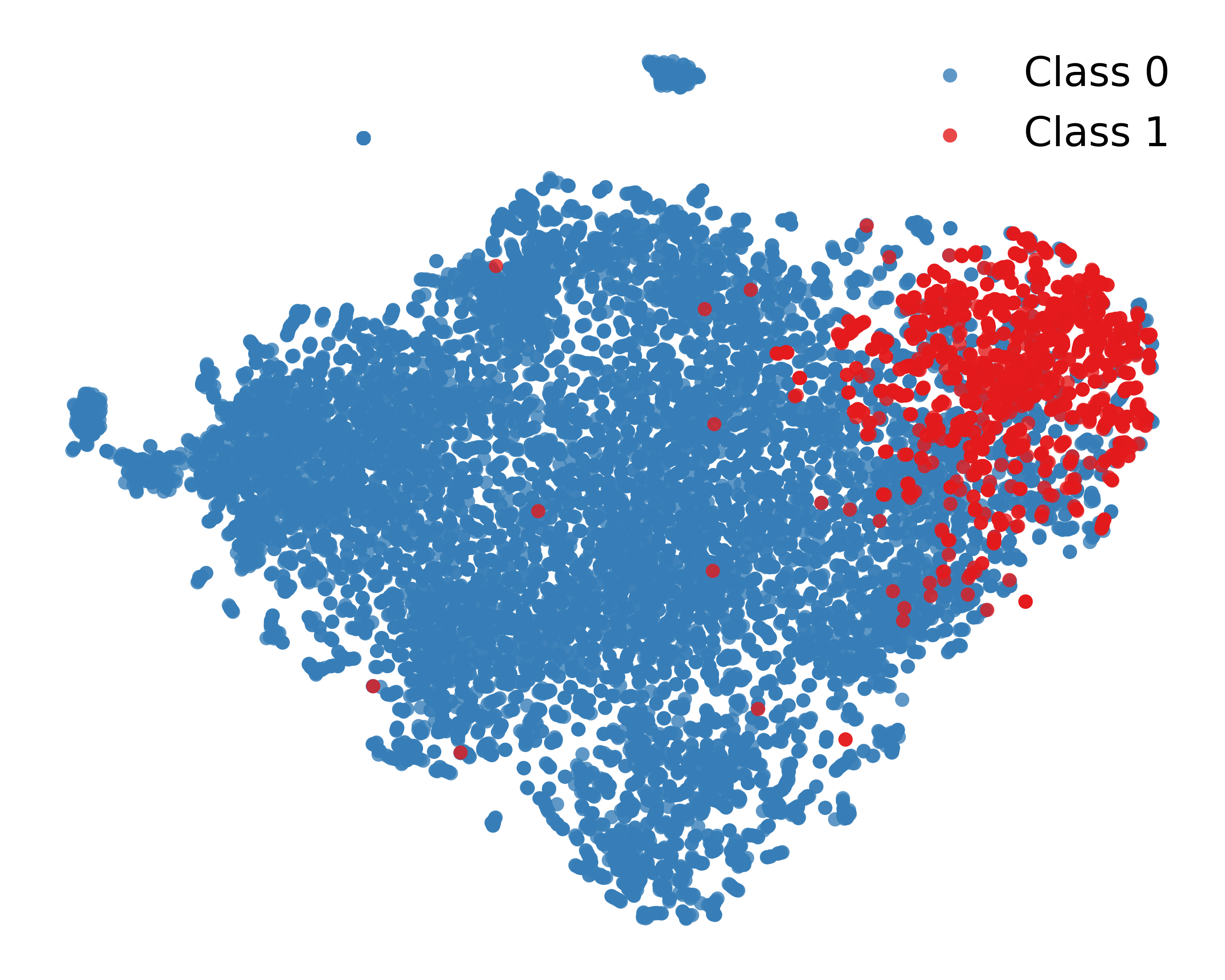}
    \caption{t-SNE}
    \label{fig:left}
    \end{subfigure}
    \caption{Visualizations of the feature embeddings extracted from the penultimate layer of FetalCLIP\textsubscript{CLS}, using PCA and t-SNE.} \label{fig:tsne}
\end{figure}

We present PCA and t-SNE visualizations of the feature embeddings extracted from FetalCLIP\textsubscript{CLS} in Figure~\ref{fig:tsne}. Both plots exhibit relatively low intra-class variance, indicating discriminative embeddings. However, noticeable inter-class overlap is observed, which may be attributed to the similarity of suboptimal frames adjacent to the optimal ones, reflecting the inherent difficulty of the dataset.

\subsection{Repurposing Segmentation Model for IQA}

As shown in Table~\ref{tab:seg_model}, FetalCLIP\textsubscript{SEG} achieves a higher F1 score and recall compared to FetalCLIP\textsubscript{CLS}. Moreover, the strong Dice score (0.7244) confirms that the segmentation decoder produces accurate masks. The results demonstrate the feasibility of repurposing a segmentation model for classification tasks through mask thresholding and show that an adapted segmentation model, supervised with pixel-wise annotations, can achieve remarkable classification performance, even surpassing a dedicated classification model on critical metrics.

\begin{table}[h]
\caption{
Model performance of FetalCLIP\textsubscript{CLS} and FetalCLIP\textsubscript{SEG}. The best average scores for each metric are \textbf{bolded}.
} \label{tab:seg_model}

\resizebox{\textwidth}{!}{
\begin{tabular}{|l|c|c|c|c|c|c|}
\hline
\textbf{Models} & \textbf{DICE}$\uparrow$ & \textbf{Accuracy}$\uparrow$ & \textbf{F1 Score}$\uparrow$ & \textbf{Precision}$\uparrow$ & \textbf{Recall}$\uparrow$ & \# \textbf{Trainable}\\
\hline
FetalCLIP\textsubscript{CLS} & / & $\textbf{0.9575} \pm 0.001$ & $0.7570 \pm 0.007$ & $\textbf{0.7782} \pm 0.034$ & $0.7397 \pm 0.041$ & 2.4 M \\
\hline
FetalCLIP\textsubscript{SEG} & $0.7244 \pm 0.007$ & $0.9543 \pm 0.001$ & $\textbf{0.7708} \pm 0.005$ & $0.6988 \pm 0.010$ & $\textbf{0.8599} \pm 0.017$ & 4.0 M \\
\hline
\end{tabular}
}
\end{table}

The primary drawback of FetalCLIP\textsubscript{SEG} is its reduced precision (0.6988), approximately 8\% lower than that of FetalCLIP\textsubscript{CLS}. These results suggest that while the segmentation-based approach is effective in identifying relevant anatomical structures, it may also produce more false positives by potentially classifying suboptimal or ambiguous frames as positive.

On the other hand, since the encoder remains frozen and the introduced decoder is lightweight, FetalCLIP\textsubscript{SEG} maintains computational efficiency and remains well-suited for deployment in environments with limited computational resources.

\begin{figure}[htbp]
    \centering
    \textbf{Ground Truth} \\
    \begin{subfigure}[b]{0.15\textwidth}
        \includegraphics[width=\linewidth]{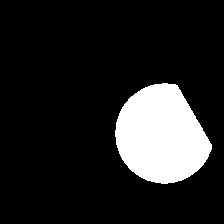}
    \end{subfigure}
    \begin{subfigure}[b]{0.15\textwidth}
        \includegraphics[width=\linewidth]{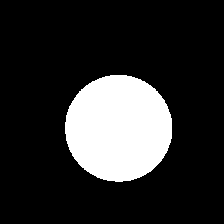}
    \end{subfigure}
    \begin{subfigure}[b]{0.15\textwidth}
        \includegraphics[width=\linewidth]{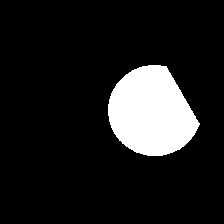}
    \end{subfigure}
    \begin{subfigure}[b]{0.15\textwidth}
        \includegraphics[width=\linewidth]{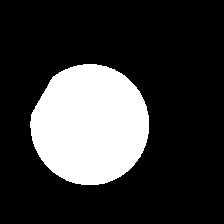}
    \end{subfigure}
    \begin{subfigure}[b]{0.15\textwidth}
        \includegraphics[width=\linewidth]{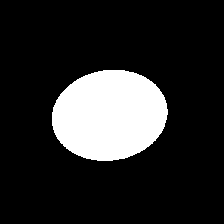}
    \end{subfigure}
    \begin{subfigure}[b]{0.15\textwidth}
        \includegraphics[width=\linewidth]{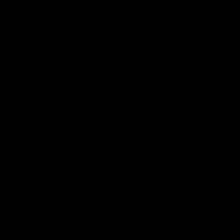}
    \end{subfigure}

    \textbf{Prediction} \\
    \begin{subfigure}[b]{0.15\textwidth}
        \includegraphics[width=\linewidth]{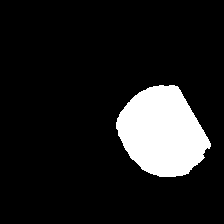}
    \end{subfigure}
    \begin{subfigure}[b]{0.15\textwidth}
        \includegraphics[width=\linewidth]{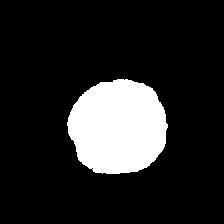}
    \end{subfigure}
    \begin{subfigure}[b]{0.15\textwidth}
        \includegraphics[width=\linewidth]{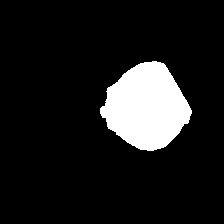}
    \end{subfigure}
    \begin{subfigure}[b]{0.15\textwidth}
        \includegraphics[width=\linewidth]{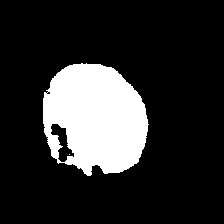}
    \end{subfigure}
    \begin{subfigure}[b]{0.15\textwidth}
        \includegraphics[width=\linewidth]{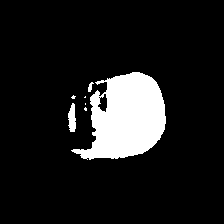}
    \end{subfigure}
    \begin{subfigure}[b]{0.15\textwidth}
        \includegraphics[width=\linewidth]{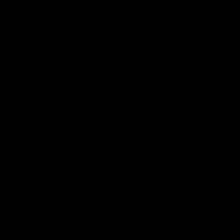}
    \end{subfigure}
    \caption{Each vertical pair displays the ground truth annotation (top) and its corresponding predicted segmentation mask (bottom). In the rightmost example, both the ground truth and prediction contain no mask, indicating a correctly segmented frame.}
    \label{fig:seg_samples}
\end{figure}

Figure~\ref{fig:seg_samples} presents six pairs of predicted segmentation masks along with their corresponding ground truth annotations, illustrating FetalCLIP\textsubscript{SEG}'s ability to accurately localize the abdominal region in fetal ultrasound frames.

\section{Conclusion}

In this work, we investigated the application of FetalCLIP for fetal ultrasound IQA, aiming to assist less-experienced sonographers in identifying suitable frames for abdominal circumference measurement from blind sweeps.

Our results demonstrate that FetalCLIP produces more effective ultrasound-specific embeddings than several of the most commonly used foundation models. FetalCLIP\textsubscript{CLS} consistently outperforms CNN and Transformer baselines in both accuracy and F1 score while maintaining competitive precision and recall with low computational overhead. Additionally, incorporating mask supervision in FetalCLIP\textsubscript{SEG} yields an improved F1 score and recall for optimal frame selection, albeit at the cost of precision.

Our study highlights that ultrasound-specific foundation models can improve diagnostic accuracy while being deployable in a resource-efficient way. Future work may extend this framework to additional fetal biometry tasks and incorporate datasets spanning a broader range of difficulty levels, facilitating more diverse and robust analyses aimed at improving prenatal care in low-resource settings.

\bibliographystyle{splncs04}
\bibliography{mybibliography}

\appendix

\section{Ablation Study}
\subsection{Selection of Fine-Tuning Strategies}
\label{sec:ablation}
We evaluate three fine-tuning strategies, including linear probing, full-parameter fine-tuning, and LoRA, using DenseNet and ViT as representative CNN- and Transformer-based models respectively.

\begin{table}[ht]
\centering
\caption{Comparison of fine-tuning strategies for DenseNet and ViT. LP and FP denote linear probing and full-parameter fine-tuning respectively. The best average scores for each metric per model are \textbf{bolded}.}

\label{tab:ablation}
\resizebox{\textwidth}{!}{
\begin{tabular}{|l|c|c|c|c|c|c|}
\hline
\textbf{Model} & \textbf{Strategy} & \textbf{Accuracy}$\uparrow$ & \textbf{F1 Score}$\uparrow$ & \textbf{Precision}$\uparrow$ & \textbf{Recall}$\uparrow$ & \textbf{\# Trainable} \\
\hline
\multirow{2}{*}{DenseNet} & LP & $0.9187 \pm 0.000$ & $0.3569 \pm 0.018$ & $0.6113 \pm 0.010$ & $0.2525 \pm 0.019$ & 1.0 K \\
& FP & $\textbf{0.9516} \pm 0.002$ & $\textbf{0.7024} \pm 0.028$ & $\textbf{0.7805} \pm 0.026$ & $\textbf{0.6420} \pm 0.059$ & 7.0 M \\
\hline
\multirow{3}{*}{ViT\textsubscript{400M}} & LP & $0.9339 \pm 0.001$ & $0.5273 \pm 0.009$ & $0.7319 \pm 0.007$ & $0.4123 \pm 0.013$ & 1.0 K \\
& FP & $0.9105 \pm 0.000$ & $0.0000 \pm 0.000$ & $0.0000 \pm 0.000$ & $0.0000 \pm 0.000$ & 303 M \\
& LoRA & $\textbf{0.9560} \pm 0.003$ & $\textbf{0.7506} \pm 0.019$ & $\textbf{0.7657} \pm 0.042$ & $\textbf{0.7417} \pm 0.067$ & 2.4 M \\
\hline
ViT\textsubscript{small} & FP & $0.9360 \pm 0.005$ & $0.5814 \pm 0.059$ & $0.7002 \pm 0.039$ & $0.5052 \pm 0.097$ & 21.7 M \\
\hline
\end{tabular}
}
\end{table}

As shown in Table~\ref{tab:ablation}, full-parameter fine-tuning outperforms linear probing for DenseNet. However, full-parameter fine-tuning of ViT\textsubscript{400M} yields zero positive predictions, likely due to its overwhelmingly large number of trainable parameters. In contrast, LoRA uses a small number of parameters and outperforms the other two strategies.

We conduct a simple experiment to verify whether zero positive predictions result from a mismatch between model and data sizes. Table~\ref{tab:ablation} shows that full-parameter fine-tuning of a smaller ViT yields marginally acceptable performance. Based on these results, we adopt full-parameter fine-tuning for CNN-based models and LoRA for Transformer-based models.

\subsection{Domain Relevance Matters in Foundation Models}

\begin{table}[ht]
\centering
\caption{Model performance of ViT models pretrained on generic image-text datasets at varying scales and FetalCLIP\textsubscript{CLS}. The best average scores per metric are \textbf{bolded}.}
\label{tab:pretrain_scale}
\resizebox{\textwidth}{!}{
\begin{tabular}{|l|c|c|c|c|c|c|}
\hline
\textbf{Model} & \textbf{Accuracy}$\uparrow$ & \textbf{F1 Score}$\uparrow$ & \textbf{Precision}$\uparrow$ & \textbf{Recall}$\uparrow$ & \textbf{\# Trainable} \\
\hline
ViT\textsubscript{400M} & $0.9560 \pm 0.003$ & $0.7506 \pm 0.019$ & $0.7657 \pm 0.042$ & $\textbf{0.7417} \pm 0.067$ & 2.4 M \\
ViT\textsubscript{2B} & $0.9555 \pm 0.002$ & $0.7429 \pm 0.011$ & $0.7687 \pm 0.017$ & $0.7196 \pm 0.026$ & 2.4 M \\

FetalCLIP\textsubscript{CLS} & $\textbf{0.9575} \pm 0.001$ & $\textbf{0.7570} \pm 0.007$ & $\textbf{0.7782} \pm 0.034$ & $0.7397 \pm 0.041$ & 2.4 M \\
\hline
\end{tabular}
}
\end{table}

Table~\ref{tab:model_performance} shows that ViT\textsubscript{400M} underperforms FetalCLIP\textsubscript{CLS} in accuracy, F1 score, and precision, despite FetalCLIP\textsubscript{CLS} being trained on only 210,000 domain-specific samples. To examine whether scaling up the generic pretraining data could bridge this performance gap, we evaluate a ViT model sharing the same architecture as ViT\textsubscript{400M}, pretrained on the LAION-2B dataset containing 2 billion image-text pairs, and present the results in Table~\ref{tab:pretrain_scale}. Notably, FetalCLIP\textsubscript{CLS} still outperforms ViT\textsubscript{2B} even though the latter is pretrained on a dataset 10,000 times larger.

While large quantities of training data are key to the generalizability of large pretrained models, our results show that comparable performance can be achieved using models pretrained on fewer but domain-specific data, highlighting the importance of domain relevance in foundation models.

\end{document}